%% file: ms.tex
\documentclass[10pt,twocolumn,letterpaper]{article}

\usepackage{cvpr}              % To produce the CAMERA-READY version

\usepackage{times}
\usepackage{epsfig}
\usepackage{graphicx}
\usepackage{amsmath}
\usepackage{amssymb}

\usepackage{enumitem}
\usepackage[ruled]{algorithm2e}
\usepackage{array}
\usepackage{multirow}
\usepackage{color}
\usepackage{cases}
\usepackage[export]{adjustbox}
\usepackage[dvipsnames]{xcolor}
\usepackage{textpos}
\usepackage{comment}
\usepackage{textcomp}
\usepackage{listings}
\usepackage{bibentry}

\usepackage{rotating}

\usepackage{booktabs}

\usepackage[pagebackref,breaklinks,colorlinks]{hyperref}

\usepackage[capitalize]{cleveref}
\crefname{section}{Sec.}{Secs.}
\Crefname{section}{Section}{Sections}
\Crefname{table}{Table}{Tables}
\crefname{table}{Tab.}{Tabs.}

\global\long\def\comma{\enspace\mbox{,}}%
\global\long\def\dot{\enspace\mbox{.}} %

\def\cvprPaperID{5} % *** Enter the CVPR Paper ID here
\def\confName{CVsports}
\def\confYear{2022}

\begin{document}

%%%%%%%%% TITLE - PLEASE UPDATE
\title{Semi-Supervised Training to Improve Player and Ball Detection in Soccer}
% Unlabeled data to improve detection in soccer
%%%% Your unlabeled soccer games just need to be pseudo-labeled.
% How to use unlabeled data to improve detection in soccer
% Unlabeled data are not useless for detection in soccer
% Marc: mes propositions
% 1. Universal Semi-Supervised Training to Improve Player and Ball Detection for Sports Analysis (premier choix !!!)
% 1.bis Universal Semi-Supervised Training to Improve Player and Ball Detection in Sport 
% 2. Universal Semi-Supervised Training to Improve Player and Ball Detection in Soccer

\author{Renaud Vandeghen$^*$\\
{\small University of Li\`ege}
% For a paper whose authors are all at the same institution,
% omit the following lines up until the closing ``}''.
% Additional authors and addresses can be added with ``\and'',
% just like the second author.
% To save space, use either the email address or home page, not both
\and
Anthony Cioppa$^*$\\
{\small University of Li\`ege}
\and
Marc Van Droogenbroeck\\
{\small University of Li\`ege}\\
}
\maketitle

\newcommand\blfootnote[1]{%
  \begingroup
  \renewcommand\thefootnote{}\footnote{#1}%
  \addtocounter{footnote}{-1}%
  \endgroup
}
\blfootnote{\textbf{(*)} Denotes equal contributions. Contacts: r.vandeghen@uliege.be and anthony.cioppa@uliege.be.}

\thispagestyle{empty}

% Notations
\newcommand{\teacher}{\mathcal{T}}
\newcommand{\student}{\mathcal{S}}
\newcommand{\finetune}{\mathcal{F}}
\newcommand{\threshold}{\tau}
\newcommand{\imageset}[1]{\mathcal{D}_{#1}}
\newcommand{\red}[1]{\textcolor{red}{#1}}

% Custom useful commands
\newcommand{\mysection}[1]{\vspace{2pt}\noindent\textbf{#1}}
\newcommand{\Table}[1]{Table~\ref{tab:#1}}
\newcommand{\Figure}[1]{Figure~\ref{fig:#1}}
\newcommand{\Equation}[1]{Equation~\eqref{eq:#1}}
\newcommand{\Equations}[2]{Equations \eqref{eq:#1} and \eqref{eq:#2}}
\newcommand{\Section}[1]{Section~\ref{sec:#1}}
\newcommand{\SoccerNet}{SoccerNet~\cite{Giancola_2018_CVPR_Workshops}\xspace}
\newcommand{\ActivityNet}{ActivityNet~\cite{caba2015activitynet}\xspace}

\newcommand{\TODO}[1]{\textcolor{red}{[TODO:#1]}}

\definecolor{myred}[a=.5]{RGB}{215,25,28} 
\definecolor{myorange}[a=.5]{RGB}{253,174,97}
\definecolor{anthoblue}[a=.5]{RGB}{31,119,180}
\definecolor{anthoorange}[a=.5]{RGB}{255,127,14}
\definecolor{anthogreen}[a=.5]{RGB}{0,150,0}
\definecolor{anthored}[a=.5]{RGB}{150,0,0}
\definecolor{anthobrown}[a=.5]{RGB}{153,76,0}
\definecolor{mygreen}[a=.5]{RGB}{166,217,106} 
\definecolor{mygray}[a=.5]{gray}{0.57}

\definecolor{newanthogreen}[a=.5]{RGB}{101,140,49}
\definecolor{newanthored}[a=.5]{RGB}{191,0,0}
\definecolor{newanthoblue}[a=.5]{RGB}{0,127,255}
\definecolor{newanthogray}[a=.5]{RGB}{76,76,76}

\definecolor{newanthoorangespotting}[a=.5]{RGB}{227,140,16}
\definecolor{newanthobluespotting}[a=.5]{RGB}{31,119,180}
\definecolor{newanthogreenspotting}[a=.5]{RGB}{44,160,44}

\definecolor{newanthoredreplay}[a=.5]{RGB}{183,27,27}
\definecolor{newanthopinkreplay}[a=.5]{RGB}{217,118,213}

\definecolor{newjacobblue}[a=.5]{RGB}{76,114,176}
\definecolor{newjacoborange}[a=.5]{RGB}{221,132,82}

\newcommand{\whitebox}{\hfill\textcolor{white}{\rule[1mm]{1.8mm}{2.8mm}}\hfill}
\newcommand{\redbox}{\hfill\textcolor{myred}{\rule[1mm]{1.8mm}{2.8mm}}\hfill}
\newcommand{\orangebox}{\hfill\textcolor{myorange}{\rule[1mm]{1.8mm}{2.8mm}}\hfill}
\newcommand{\greenbox}{\hfill\textcolor{mygreen}{\rule[1mm]{1.8mm}{2.8mm}}\hfill}
\newcommand{\graybox}{\hfill\textcolor{mygray}{\rule[1mm]{1.8mm}{2.8mm}}\hfill}
\newcommand{\BG}[1]{\textbf{{\color{red}[BG: #1]}}}

%%%%%%%%% ABSTRACT
\input{sections/0_abstract.tex}

%%%%%%%%% BODY TEXT
\input{sections/1_introduction.tex}

% \newpage
\input{sections/2_stateart.tex}

% \newpage
\input{sections/3_method.tex}

% \newpage
\input{sections/4_experiments.tex}

% \newpage
\input{sections/5_conclusion.tex}
\clearpage

%%%%%%%%% REFERENCES
{\small

\input{ms.bbl}
%\bibliographystyle{ieee_fullname}
%\bibliography{abbreviation-short,dataset,labo,learning,library,soccer,sports,stod,vision,new-refs}
}

\end{document}

%% file: sections/0_abstract.tex
% "The maximum size of the abstract is 4000 characters."
\begin{abstract}
Accurate player and ball detection has become increasingly important in recent years for sport analytics. 
As most state-of-the-art methods rely on training deep learning networks in a supervised fashion, they require huge amounts of annotated data, which are rarely available.
In this paper, we present a novel generic semi-supervised method to train a network based on a labeled image dataset by leveraging a large unlabeled dataset of soccer broadcast videos.
More precisely, we design a teacher-student approach in which the teacher produces surrogate annotations on the unlabeled data to be used later for training a student which has the same architecture as the teacher.
Furthermore, we introduce three training loss parametrizations that allow the student to doubt the predictions of the teacher during training depending on the proposal confidence score.
We show that including unlabeled data in the training process allows to substantially improve the performances of the detection network trained only on the labeled data. 
Finally, we provide a thorough performance study including different proportions of labeled and unlabeled data, and establish the first benchmark on the new SoccerNet-v3 detection task, with an mAP of $52.3\%$. Our code is available at [\url{https://github.com/rvandeghen/SST}].
\end{abstract}

%% file: sections/1_introduction.tex
\section{Introduction}
\label{sec:Intro}

\begin{figure}
    \centering
    \includegraphics[width=\linewidth]{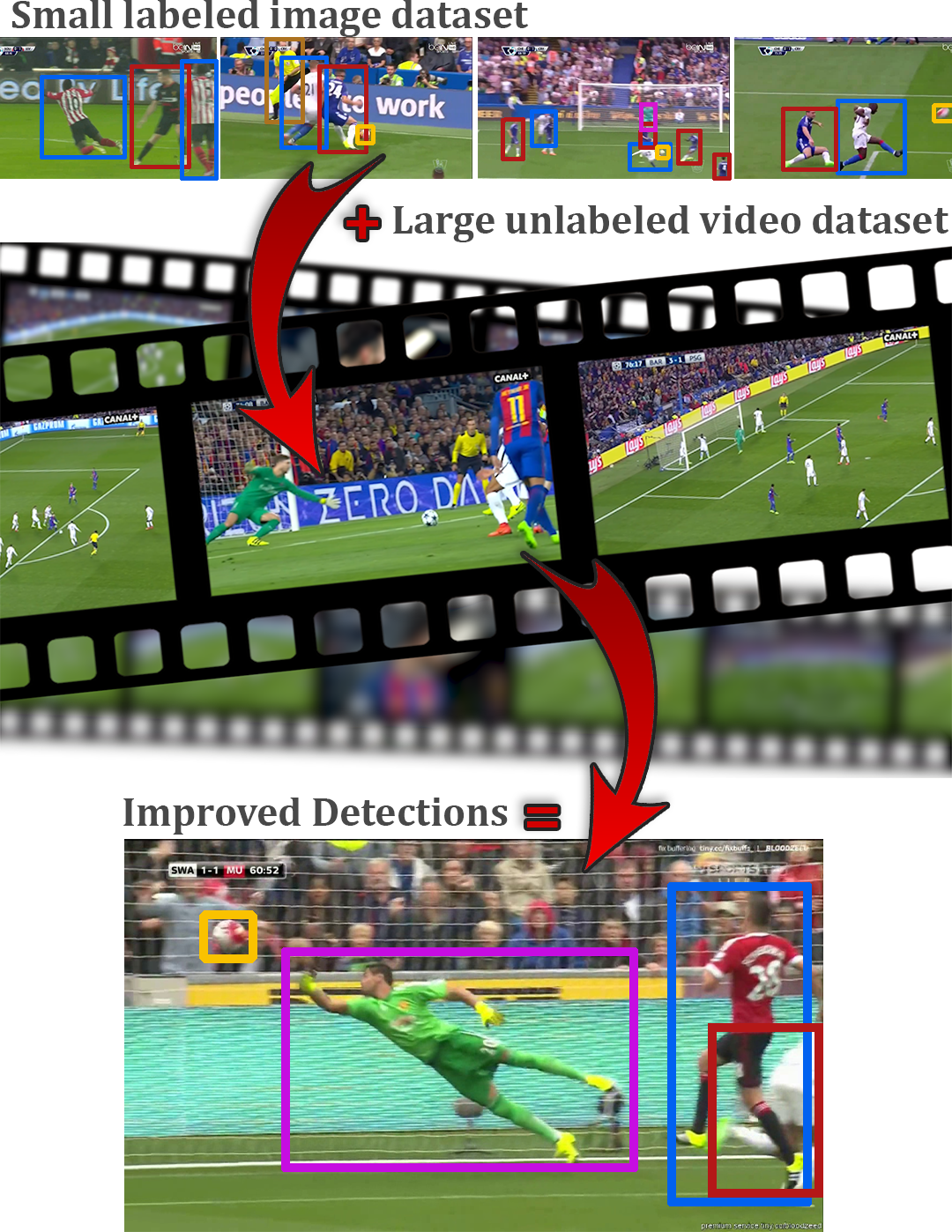}
    \caption{\textbf{Overview.} Given a small labeled image dataset for object detection in soccer such as the players, the ball, or the referees, we leverage a large unlabeled dataset of soccer broadcast videos for training an object detector in a semi-supervised fashion. Our training technique allows us to significantly improve the performance of the object detector for the targeted soccer application.}
    \label{fig:graphical_abstract}
\end{figure}

Sports analytics has been steadily growing over the last decade~\cite{DTAI2019WhySports}, pushed by the development of advanced artificial intelligence and computer tools. 
Last year, the market was estimated at more than $1$ billion dollars, with most indicators pointing out a growth by $500\%$ within the next 5-10 years~\cite{Mordorintelligence2022Sports,Gough2021Market}. 
Therefore, sports analytics will become even more central for the sports industry in the coming years. 
Some companies already offer analytics services to clubs with the purpose to improve their playing performances and ascend the championship ladder, thus generating more revenues from ticket sales, advertisements and merchandising.
%Coaches may also improve their game strategy and analyze individual player performances in prevision of the next transfer market.

Nowadays, most sports analytics products either rely on manual inspection, which has a heavy cost in terms of manpower, or more recently on automated analysis systems based on artificial intelligence and computer vision techniques.
The first step of automated systems often relies on accurately retrieving the players and the ball, which are the key elements to grasp the course of the game.
From this information, deeper analyses may be performed such as tracking the players to extract individual speed performance, estimating the field coverage by a defending team to unveil potential weaknesses, or analyze critical pass decisions.
All these are powerful indicators of an individual's performance, and game strategy analyses may reveal the strengths and weaknesses of opponent or one's own team.
Accurately detecting the players and the ball is therefore crucial since analyses rely on these preliminary results.

Over the past few years, artificial intelligence techniques have surpassed their hand-crafted features algorithms counterparts in many areas including player and ball detection in sports.
Even though many deep learning detection networks are publicly available for sports companies and researchers, they are often trained on generic data that are not specifically tailored for each sport.
The domain gap between the training dataset and the targeted application often results in performances lower than expected, which is why training, or at least fine-tuning, on sport specific data is often required.
However, this may require huge amounts of data, which can be costly to annotate and cannot be transferred from one sport to another.
Furthermore, some recent works showed that training the network on sport, and even stadium or team, specific data allows to substantially improve the performance of those networks~\cite{Cioppa2019ARTHuS}.

In this paper, we present a novel generic semi-supervised method for training an object detector with few annotated soccer images, by leveraging a large unlabeled dataset of soccer broadcast videos as illustrated in Figure~\ref{fig:graphical_abstract}.
More specifically, we develop an iterative teacher-student training approach with three different training loss parametrizations for the student, which may doubt the detections performed by the teacher based on their confidence score.
We show that including unlabeled data in the training process allows to substantially increase the performances of the detection network on unseen soccer games.
Specifically, we provide a complete performance study for different proportions of labeled and unlabeled data, and establish the first benchmark for the detection task on the new SoccerNet-v3~\cite{Cioppa2022SoccerNetv3} 
dataset.
It is important to note that the presented ideas and achievements do not rely on any data knowledge about soccer, nor on the network architecture. 
%Therefore, our method may be applicable to any other domain with the same data availability, which is particularly the case for other sports.
Therefore, our method is applicable to any other sport or domain, characterized by a low amount of annotated data and a large dataset of unlabeled data, and for any detection network.

%%%%%%%% OUR CONTRIBUTION
\mysection{Contributions.} We summarize our contributions as follows. 
\textbf{(i)} We propose a novel semi-supervised method for training a player and ball detection network in soccer games with a teacher-student approach.
\textbf{(ii)} We introduce three loss parametrizations for training the student with the objective to doubt detections performed by the teacher based on their confidence scores.
\textbf{(iii)} We establish the first detection benchmark on the new SoccerNet-v3 dataset.

%% file: sections/2_stateart.tex
\section{Related Work}
\label{sec:SOTA}

\mysection{Object detection in sport analytics.}
Object detection has been massively studied in the context of sports analytics as it provides a strong basis for further analyses techniques~\cite{Thomas2017Computer}. 
Even though the first detection algorithms used background subtraction to detect players\cite{Archana2015AnEfficient,Rao2015ANovel}, they have been quickly overthrown by deep learning networks such as convolutional neural networks (CNN).
For instance, the authors of~\cite{Sah2018Evaluation} use a shallow CNN to detect players on a hockey field with different image representations.
Other methods rely on pre-trained networks such as Mask R-CNN\cite{He2017Mask,Pobar2018MaskRCNN,Yang20183DMultiview}. 
Recently, Cioppa \etal\cite{Cioppa2020Multimodal} proposed a cross-modality online distillation method for player detection and counting on low budget stadium. 
Liu \textit{et  al.}~\cite{Liu2021DetectingAM} developed a method to detect players and automatically match them with object such as hockey players and their stick.

Some other works use detection as a first step for various downstream tasks such as improving action spotting using camera calibration and player localization~\cite{Cioppa2021Camera}, player and ball tracking~\cite{Hurault2020SelfSupervised,Manafifard2017ASurvey, Kamble2019Deep}, or to model pass feasibility~\cite{Sangesa2020UsingPB}.

In order to train deep learning networks, the AI for sports community can count on a large variety of datasets for sports analytics. SoccerNet~\cite{Giancola2018SoccerNet} and SoccerNet-v2~\cite{Deliege2021SoccerNetv2} propose 500 complete broadcast soccer games with annotated action events, camera cuts and classes, and replay information. A complementary dataset with spatio-temporal event annotations focusing on player statistical analyses was released by Pappalardo \etal~\cite{Pappalardo2019Apublic}. Yu \etal~\cite{Yu2018Comprehensive} and SoccerDB, published by Jiang \etal~\cite{Jiang2020SoccerDB}, provide annotations for more than $200$ soccer games with player bounding boxes and shot transitions. Lately, SoccerNet-v3
~\cite{Cioppa2022SoccerNetv3} 
was released, providing manual bounding box annotations for player and other objects of interest such as the ball, the lines, and the goal, with extra annotations such as jersey numbers and re-identification of players across multiple views.

\mysection{Object detection in general.}
Together with image classification, object detection is among the most studied task in computer vision.
Many object detection architectures have been developed in the past few years thanks to the availability of large-scale datasets such as Pascal VOC~\cite{Everingham2010PascalVOC} or MS COCO~\cite{Lin2014Microsoft}.
Usually, object detectors come into one of two main flavors: two-stage detectors~\cite{Girshick2015Fast, Ren2017Faster, He2017Mask, Girshick2014Rich, Lin2017Feature}, and one-stage detectors~\cite{Liu2016SSD, Redmon2016YOLO, Redmon2018YOLOv3, Lin2017Focal, Tan2020EfficientDet}.
For two-stage detectors, a proposal module is used to propose regions of interest where potential object candidates are likely to be located, for example with a region proposal network such as in Faster R-CNN~\cite{Ren2017Faster}. The proposals are later refined in a second module, where a class is associated with each predicted bounding box.
One-stage detectors operate differently, and directly output the bounding boxes with their classes, leading to faster inference, but often at the price of a lower accuracy compared to their two-stage counterparts.
For these reasons, in this work, we will focus on the two-stage Faster-RCNN~\cite{Ren2017Faster} architecture, which is widely used in semi-supervised object detection.
Note however that our method is applicable regardless of the network architecture.

\mysection{Semi-supervised object detection.}
Following the successes of semi-supervised methods achieved for image classification~\cite{Berthelot2019MixMatch, Berthelot2020ReMixMatch, Pham2021Meta, Sohn2020FixMatch, Xie2020SelfTraining, Zoph2020Rethinking}, many semi-supervised learning methods for object detection have been developed over the past few years.
In $2019$, Jeong \etal~\cite{Jeong2019Consistency} proposed a consistency method for the detections made for an image and its horizontally flipped version.
More recently, Sohn \etal~\cite{Sohn2020Simple} designed a teacher-student approach~\cite{Li2020Improving, Liu2021Unbiased, Tang2021Proposal, Zoph2020Rethinking, Xu2021Soft}, where the teacher model is trained with labeled data in a supervised manner, and used to produce pseudo-labels on the unlabeled data.
These pseudo-labels, along with the labeled data, are then used to train the student model, leading to better performances.
This teacher-student approach relies on a selection mechanism to include or reject pseudo-labels, which is often performed by comparing their confidence score to a threshold.
However, determining the appropriate threshold value is an arduous process as it is prone to generate noise, resulting in false positives or false negatives.
Therefore, authors have promoted different learning strategies for the student, including Unbiased Teacher~\cite{Liu2021Unbiased}, which addresses the bias issue regarding the dominant classes with a weighted focal loss~\cite{Lin2017Focal} for the classification head, and Soft Teacher~\cite{Xu2021Soft}, which uses a confidence score for each pseudo-label to weight the background classification loss.
In this paper, we present a weighting strategy on the foreground boxes rather than the background ones, with a doubt mechanism based on the confidence score of the pseudo-labels.

%% file: sections/3_method.tex
\section{Method}
\label{sec:Method}

\begin{figure*}
    \centering
    \includegraphics[width=\textwidth]{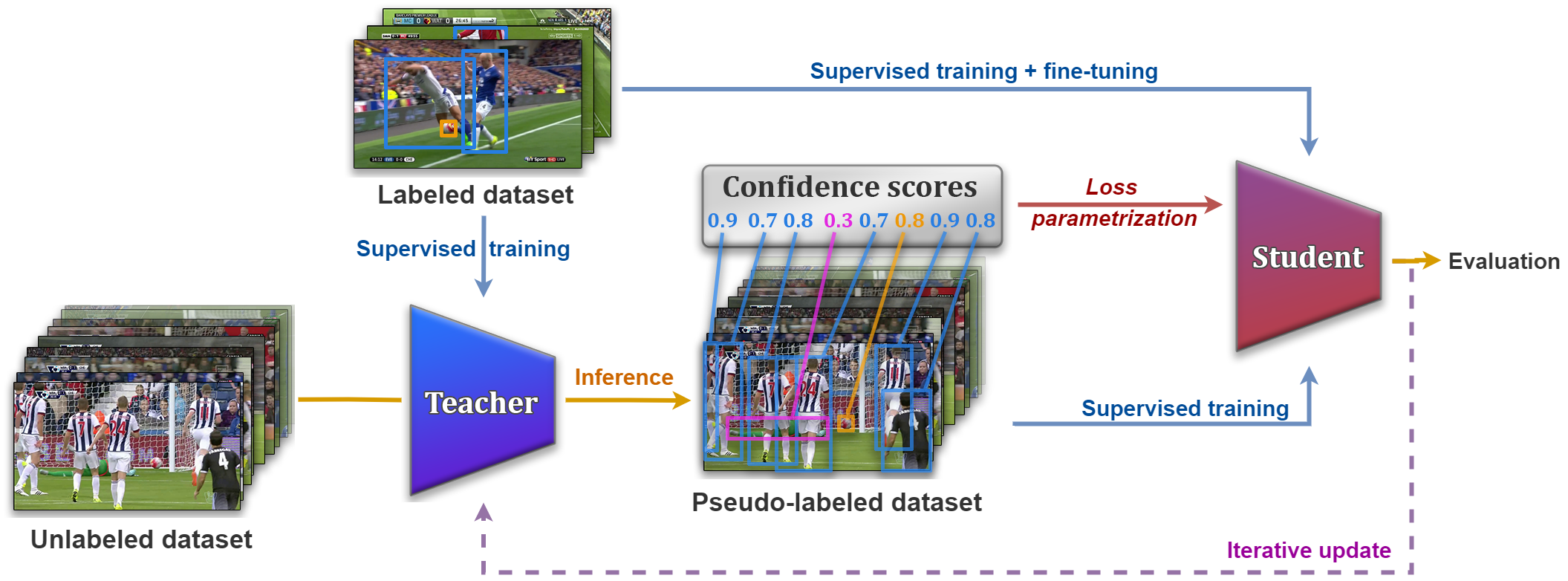}
    \caption{\textbf{Overview of our semi-supervised training method for player and ball detection.} We first train a teacher network on a labeled dataset in a fully supervised fashion. Then, we use the trained teacher to produce pseudo-labels on the unlabeled dataset. This creates a first pseudo-labeled dataset, with a confidence score for each prediction. The labeled and pseudo-labeled datasets are then used to train a student network, whose training loss is parameterized based on the confidence score with one of the three parametrization introduced in this paper. This allows the student to doubt unsure proposals by the teacher and achieve good performances on the test dataset. At the end of the training, a final fine-tuning phase is performed with the labeled data, and the student becomes the new teacher for the next iteration.}
    \label{fig:method}
\end{figure*}

\mysection{Problem statement.}
We leverage the availability of unlabeled data to improve the detection performance as follows.
Given a model tailored for a detection task on images, and trained with a dataset $\imageset{l}$ comprising $N_l$ labeled images, we make use of a dataset $\imageset{u}$ comprising $N_u$ unlabeled images to increase the detection performance of the model; annotations of a labeled image consist in the bounding boxes and classes for all objects contained in it.

This setup is very common in artificial intelligence as datasets are extremely time-consuming and expensive to annotate. 
Therefore, only a tiny portion of the available data is usually annotated and used for training a model.
In this work, we show how to exploit unlabeled images in a semi-supervised fashion for sports analysis.
In particular, we propose a method based on a teacher-student approach, where a teacher model $\teacher$ is trained only with the labeled data, and a student model $\student$ is trained with the labeled and unlabeled, for which pseudo-labels are produced by $\teacher$.

\mysection{Iterative semi-supervised training.}
The first step of our method consists in training the teacher model $\teacher$ with a standard supervised learning technique on the labeled dataset $\imageset{l}$.
Once $\teacher$ is properly trained, we generate pseudo-labels for images of the unlabeled dataset $\imageset{u}$.
More precisely, $\teacher$ processes each image of $\imageset{u}$ and outputs the box, class and confidence score for each detected object.
To avoid multiple predictions of the same object, a classical non-maximum suppression is performed.
Let us note that, at this point, the performance of $\teacher$ corresponds to the typical case of training a model in a supervised fashion on a labeled dataset.
Hence, the performance of the first teacher $\teacher$ is the baseline for comparisons in Section~\ref{sec:Exp}.

The next step consists in training a student $\student$, which has the exact same architecture as $\teacher$, on both $\imageset{l}$ and $\imageset{u}$. 
The training is performed in a supervised fashion, identical to that of $\teacher$, but on a larger concatenated dataset (that could be seen as a dataset augmented by $\imageset{u}$).
The training loss of $\student$ is taken as the sum of two equal contributions, that is 
\begin{equation}\label{eq:total-loss}
    \mathcal{L} = \mathcal{L}_l + \mathcal{L}_u \comma
\end{equation}
with $\mathcal{L}_l$ and $\mathcal{L}_u$ corresponding to the loss on the labeled dataset and unlabeled dataset, which now contains pseudo-labels, respectively.
Once the training is stopped, we fine-tune $\student$ with $\imageset{l}$, to make sure to finalize the training on real ground-truth annotations.
While being known in the machine learning community and to the best of our knowledge, the fine-tuning step has only been used once before by Li \textit{et. al}~\cite{Li2020Improving} in a self-training method for object detection, despite being highly efficient, as shown in Section~\ref{sec:Exp}.

These two steps (generating the pseudo-labels with $\teacher$ and training $\student$) may be iterated, by considering the last student as the new teacher and re-generating the pseudo-labels on $\imageset{u}$.
Hopefully, since the prediction quality of $\student$ is expected to be higher than $\teacher$, the next pseudo-labels should be better as well and improve the training of the next student. 

Since $\teacher$ is not perfect (otherwise we could stop the training process there), $\imageset{u}$ will contain truly detected objects (true positives), but also some predictions that do not correspond to any real objects (false positives), as well as some missing objects (false negatives).
These errors in $\imageset{u}$ affect the training of $\student$, and require to find the best practical trade-off.
In the following, we propose three training loss parametrizations for the student based on the confidence score of the proposals in order to reduce the impact of potential errors.
The whole pipeline is drawn in Figure~\ref{fig:method}.

\mysection{Loss parametrization 1: single threshold.}\label{sec:soft-threshold}
A first way to alleviate false positives in the dataset consists in selecting a subset of the pseudo-labels in $\imageset{u}$ to only retain the true positive predictions and remove the false positive ones. 
This is usually done by solely keeping predictions with a confidence score higher than a given threshold $\threshold_h$.
This reduces the number of positive proposals in the $\imageset{u}$ dataset and increases the number of background proposals.
The training loss term $\mathcal{L}_l$ of Equation~\ref{eq:total-loss}, corresponding to the labeled dataset during the training of the student, can be written as:
\begin{equation}\label{eq:loss_l}
    \mathcal{L}_l = \sum_{i=1}^{N_l}\sum_j \mathcal{L}_{cls}
    %(I_i, \mathbf{b}_{i}^{j})
    + \mathcal{L}_{reg} %(I_i, \mathbf{b}_{i}^{j}) 
    \comma
\end{equation}
where $\mathcal{L}_{cls}$ and $\mathcal{L}_{reg}$ denote the classification and box regression loss respectively, and the superscript $j$ stands for the $j$th proposal for image $i$.
Likewise, the training loss on the unlabeled dataset, $\mathcal{L}_u$, %in the training of the student of Equation~\ref{eq:total-loss} 
can be written as:
\begin{equation}\label{eq:loss_u1}
    \mathcal{L}_u = \sum_{i=1}^{N_u}\sum_j \mathcal{L}_{cls}
    %(I_i, \mathbf{\bar{b}}_i^j) 
    + \mathcal{L}_{reg}
    %(I_i, \mathbf{\bar{b}}_i^j)
    \dot
\end{equation}
Recent works~\cite{Liu2021Unbiased, Sohn2020Simple, Tang2021Proposal, Xu2021Soft} have shown that using a relatively high threshold value ($\threshold \geq 0.7$) ensures pseudo-labels of high quality.
This parametrization has two effects: (1) it allows to keep predictions which are supposedly true positives, and (2) predictions boxes with low confidence score are associated to the background and therefore correctly removed.
However, the downside is that true positive predictions may also have a confidence score lower than this threshold, leading to the introduction of incorrect false negatives in the dataset.
In fact, the threshold value acts as a trade-off between precision and recall, given that lower values tend to increase the recall despite lowering the precision, whereas higher threshold values have the opposite effect.
Thus, with the choice of a high threshold value, the trade-off tends towards a higher precision, at the price of introducing false negatives.

\mysection{Loss parametrization 2: double threshold and doubt.} In order to take into account the potential false negatives, we introduce a second threshold value $\threshold_l$ separating true background predictions with a very low confidence score from the remaining predictions.
The goal of this second threshold is to create a range of confidence scores, that is $[\threshold_l;\threshold_h]$, for which we ignore whether the predictions belong to an actual objects or not.
For all predictions with a confidence score in this range, we set the loss to $0$ so that the proposals are neither used as positive nor negative examples.
This allows to introduce doubt in the training process of the student for unsure predictions of the teacher.
The training loss for $\imageset{l}$ is the same as for the first parametrization, but now for $\imageset{u}$, we modify  Equation~(\ref{eq:loss_u1}) to introduce the new doubt range:
\begin{equation}\label{eq:loss_u2}
    \mathcal{L}_u = \sum_{i=1}^{N_u}\sum_j \alpha_j\left(\mathcal{L}_{cls}
    %(I_i, \mathbf{\bar{b}}_i^j) 
    + \mathcal{L}_{reg}
    %(I_i, \mathbf{\bar{b}}_i^j) 
    \right) \comma
\end{equation}
where the term $\alpha_j$ is defined as follows:
% for labeled images and
\begin{equation}\label{eq:loss_a1}
    \alpha_j =
    \left\{
    \begin{array}{ll}
       0 & \text{if \; $\threshold_l \leq s_j < \threshold_h$},\\
       1 & \text{otherwise,}
    \end{array}
  \right.
\end{equation}
where $s_j$ is the confidence score associated to the $j$th proposal.
Thus, pseudo-labels whose confidence score lies between $\threshold_l$ and $\threshold_h$ do not contribute anymore to the loss term $\mathcal{L}_u$.
By doing so, we can increase the value of $\threshold_h$, ensuring that the positives that we introduce actually correspond to true positives regardless of false negatives introduced in the previous parametrization.
This provides more flexibility than for the first parametrization.
% Note that for the special case when $\threshold_l = \threshold_h$, we fall back on the first parametrization.

\mysection{Loss parametrization 3: double threshold and progressive doubt.} Finally, one could argue that predictions with a confidence score close to $\tau_h$ are more reliable than predictions with scores close to $\tau_l$.
Therefore, we adapt the second parametrization by introducing a doubt that decreases between the two thresholds.
This allows us to tune the uncertainty from high for predictions close to $\tau_l$, to low for predictions as their confidence score approaches $\tau_h$.
Equations~\eqref{eq:loss_l} and \eqref{eq:loss_u2} stay the same, but Equation~\eqref{eq:loss_a1} becomes:
\begin{equation}\label{eq:loss_a2}
    \alpha_j =
    \left\{
    \begin{array}{ll}
       \frac{s_j-\tau_l}{\tau_h-\tau_l} & \text{if \;  $\threshold_l \leq s_j < \threshold_h$},\\
       1 & \text{otherwise.}
    \end{array}
  \right.
\end{equation}

\begin{figure}
    \centering
    \includegraphics[width=\linewidth]{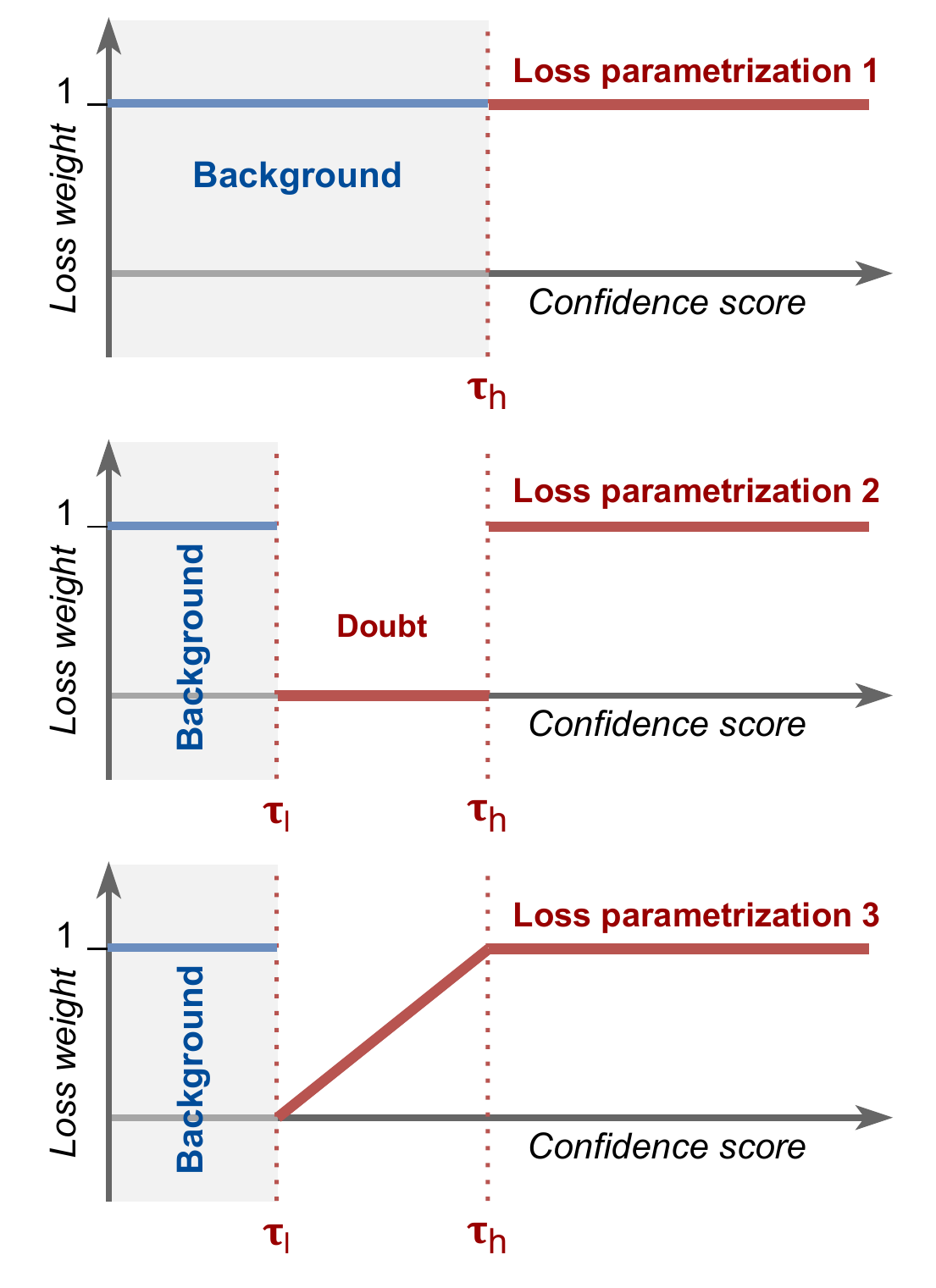}
    \caption{\textbf{Our three loss parametrizations for positive candidates.} Comparison of the evolution of the proposal loss weight (corresponding to $\alpha_j$) with respect to the prediction confidence score for our three parametrizations for positive candidates (in red). (1) Simple threshold value to discriminate between the positive proposals and the background by assigning the same loss weight to all positive samples. (2) Introduction of a second threshold to delimit a doubt zone where the loss is zeroed out. (3) Soft linear approximation for the loss weight in the doubt zone to give more importance to predictions close to $\tau_h$. Note that the loss weight is always $1$ for background proposals (in blue), regardless of the parametrization for the positive proposals.}
    \label{fig:loss_weights}
\end{figure}
The weighting term of our three parametrizations, for the loss associated with each positive proposal, is illustrated in Figure~\ref{fig:loss_weights}.
Note that for the three parametrizations, the weight loss associated with negative proposals is unchanged, regardless of the confidence score, as the background follows a different dynamic than the foreground. Indeed, it is not possible to assign a confidence score to a region without proposals from the teacher. In other terms, this means that we cannot alleviate false negatives already present in the pseudo-labeled dataset. False negative region proposals based on video analysis as in~\cite{Cioppa2020Multimodal}, and a loss parametrization for the rejected proposal ($\leq\tau_l$) may be considered. However, they are out of the scope of this paper and could be studied in a further work.

%% file: sections/4_experiments.tex
\section{Experiments}
\label{sec:Exp}

\mysection{Dataset.} 
The SoccerNet\cite{Giancola2018SoccerNet} dataset provides the largest public soccer video collection, including 550 complete broadcast games from the six most influential soccer championships in Europe. 
Recently, new annotations were released as part of SoccerNet-v3~\cite{Cioppa2022SoccerNetv3}
including $344{,}660$ human bounding boxes of players, referees, and staff, and $26{,}939$ annotations of salient objects such as the ball.
These annotations are spread across $33{,}986$ images representing salient moments in soccer with actions such as goals, cards, corners, and their replays.

We choose the training set of SoccerNet-v3 as our labeled dataset, which contains $24{,}459$ frames, its validation set to evaluate performance during training and compare the different loss parametrizations, with $4{,}797$ frames, and its test set for evaluating our final performance, with $4{,}730$ frames.
For our unlabeled set, we first retrieve the broadcast videos of the training set games of SoccerNet, which accounts for about $435$ hours of video, and extract images at 1 frame per second. 
This amounts to almost $1{,}6$ million unlabeled frames across $290$ different games, which is $64$ times more images than the labeled training set!

For the detection task, we focus on the six most important classes for soccer analysis: player, goalkeeper, main referee, side referee, staff, and ball.
This amounts to more than $250{,}000$ ground-truth bounding boxes with a highly non-uniform class distribution. % as shown in Figure $XXX$.
This dataset allows us to study our method in many cases ranging from few to many labeled and unlabeled data, with class imbalance and a wide range of object sizes, covering most practical use cases.

\mysection{Training setup.}
Both the teacher and student models are based on the same Faster R-CNN~\cite{Ren2017Faster} architecture with FPN~\cite{Lin2017Feature} and a ResNet-50~\cite{He2016DeepResidual} backbone pre-trained on ImageNet.
Therefore, these networks are composed of a first-stage region proposal network (RPN) and a second-stage detection network, each having their own classification and regression losses for training.
Regarding Equations~\eqref{eq:total-loss}, \eqref{eq:loss_l}, \eqref{eq:loss_u1} and \eqref{eq:loss_u2}, we simply equivalently consider the RPN and detection losses as described in those equations, with the total loss becoming the sum of all four losses.

For the first training phase of the teacher on the labeled dataset, we use the SGD optimizer with an initial learning rate of $0.02$, momentum of $0.9$, and a weight decay of $10^{-4}$.
We choose to evaluate our model on the validation set with the mAP ($AP_{50:95}$) metric after every epoch, which is a common metric for object detection.
If no improvement is made regarding the mAP for $5$ consecutive epochs, we reduce the learning rate by a factor of $10$.
The models are trained using $4$ GPUs with $8$ images per batch per GPU, with synchronized batch normalization layers across the different GPUs.
For both the RPN and detection modules of Faster R-CNN, we use the standard smooth L1 loss for the regression part $\mathcal{L}_{reg}$ and the cross-entropy loss for the classification part $\mathcal{L}_{cls}$.
Note that for the detection module, we also weight the classification loss for each proposal according to the class proportion in $\imageset{l}$, which is a common procedure to counter the class imbalance problem. 
Specifically, this prevents the networks from focusing too much on the most represented class such as players compared to less represented ones like the balls.
Furthermore, we use a simple data augmentation process in which we randomly apply horizontal flipping and color jittering for each training sample.
Finally, as an early stopping strategy, we cut off the training of the model if no improvement is made with respect to the mAP on the validation set for $10$ consecutive epochs or if the training reaches $200$ epochs.

Next, during the inference phase of the teacher, we process all frames of the unlabeled dataset and gather all detection with their confidence scores, localization, and classes, creating the pseudo-labeled dataset.
Afterwards, the student network is trained on both the labeled and pseudo-labeled dataset by randomly mixing the samples of both datasets.
The exact same training procedure than the one for the first teacher is used except that for each sample of the pseudo-labeled dataset, we parameterize the training loss according to one of the three techniques introduced in Section~\ref{sec:Method}.
Once the student finishes training, either by early stopping or by reaching the maximal number of epochs, we fine-tune it on the labeled dataset only.

Finally, the student network is evaluated and becomes the new teacher network for the next iteration.
The pseudo-labeled dataset is re-computed with this new teacher and a new student is trained following the above procedure.

\mysection{Quantitative results.}
\input{tables/table1.tex}
We evaluate our method on increasing labeled dataset sizes to study scenarios ranging from very few to lots of annotated data.
In particular, we select the following sizes: $1\%, 5\%, 10\%$, and $100\%$ of $\imageset{l}$, which corresponds to $3$, $14$, $29$ and $290$ games ($193$, $1{,}196$, $2{,}475$, and $24{,}459$ frames respectively).
The sampling is operated at the match level rather than at the frame level to stay close to a real-world application in which new data comes from a whole game.
For the unlabeled dataset, it is unfortunately too slow to train the model on the whole unlabeled dataset for each setup.
Therefore, for most of our experiments, we sample $10$ extra matches, not belonging to the labeled matches, which represents around $55{,}000$ frames.
Nevertheless, we evaluate our method once on the entire labeled and unlabeled datasets (corresponding to $1{,}596{,}387$ frames) for the best set of parameters found on the restricted unlabeled dataset, which defines the first detection benchmark on the SoccerNet-v3 dataset.
Those choices follow the recommendations of Oliver \etal~\cite{Oliver2018Realistic} regarding the evaluation of semi-supervised learning methods.

For each labeled dataset size and each loss parametrization, we optimize the threshold values $\threshold_l$ and $\threshold_h$ using a grid search strategy on the validation set according to good practice in semi-supervised learning.
A complete ablation study of these parameters is presented in the next subsection.
The results for the fine-tuned student models after the first iteration may be found in Table~\ref{tab:main-results}.
As can be seen, the optimal threshold values $\threshold_l$ and $\threshold_h$ are quite high for the three loss parametrizations, indicating that we select predictions for which the teacher is extremely confident.
Furthermore, for all dataset sizes, each parametrization systematically outperforms the teacher, which is the baseline corresponding to a strictly supervised approach.
We can also see that the second and third parametrizations have comparable results, but operate better than the first parametrization with a single threshold.
This indicates that doubt introduced by those parametrizations is beneficial for training the student.

Then, we evaluate only once our method trained with the entire labeled and unlabeled datasets on the test set, choosing the best performing loss parametrization and thresholds based on the previous experiments with the restricted unlabeled dataset.
As can be seen in Table~\ref{tab:main-results}, the best performing method on $100\%$ of the training data with $10$ extra games is obtained with the third parametrization and threshold values of $\threshold_l=0.9$ and $\threshold_h=1$.
Therefore, we train a student model on the whole labeled and unlabeled dataset with those parameters as well.
Since this experiment has a high training time, a single iteration is performed.
We achieve an mAP of \textbf{$52.3\%$} with the fine-tuned student, improving the performance of the teacher by $1.3\%$, which is slightly better than with $10$ extra unlabeled games ($52.0\%$ on the test set).
This shows that our method improves the detection performance compared with fully supervised methods, especially when considering few annotated data and that more unlabeled data leads to greater improvements.
\begin{figure}
    \centering
    \includegraphics[width=\linewidth]{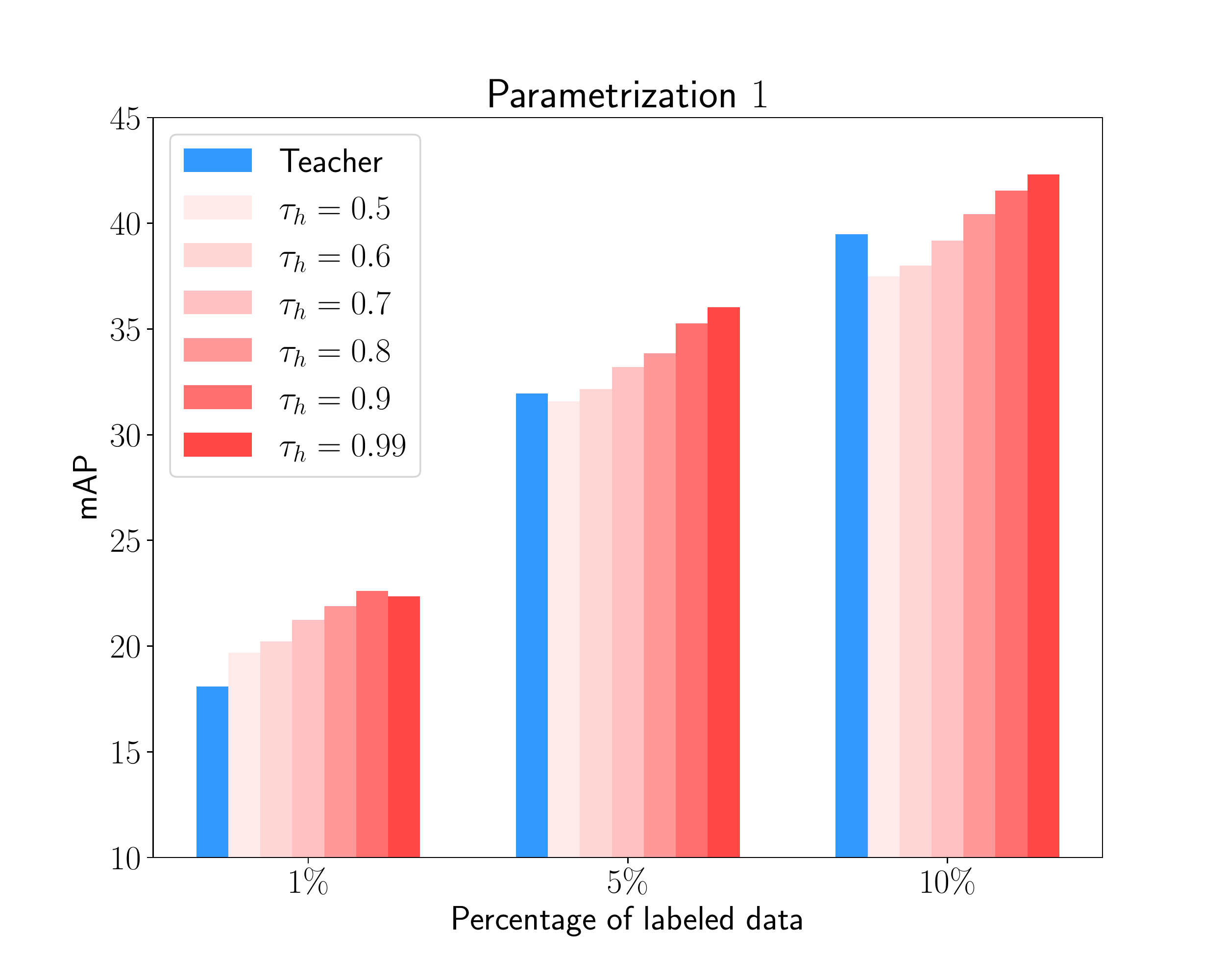}
    \caption{\textbf{Optimal threshold value for the first parametrization.} Comparison of the performance of the first parametrization for different threshold values $\threshold_h$ on various labeled dataset sizes, with $10$ extra unlabeled games. The performance of the student increases with the threshold value indicating that only predictions for which the teacher is certain should be considered. Also, the student manages to surpass the teacher for each dataset size.}
    \label{fig:p1_ablation}
\end{figure}

\mysection{Ablation study.}
In this analysis, we start by reviewing the effect of fine-tuning the student, then we propose a thorough study of $\threshold_l$ and $\threshold_h$ for our three loss parametrizations, and finally, we explore the further gain one can expect when considering multiple iterations of our method.

First, we discuss the benefit of fine-tuning the student on $\imageset{l}$ at the end of the training process.
Table~\ref{tab:finetuning} shows the performance of the student before and after fine-tuning for each dataset sizes on the validation set (the results on the right of the arrow are the ones of Table~\ref{tab:main-results}).
As can be seen, fine-tuning allows to significantly improve the performance no matter the parametrization or the labeled dataset size.
For this reason, in this ablation study, we only consider the performance \emph{before fine-tuning} as this step takes consequent computation time and that the important observations can be made on the differences between the performances rather than their absolute values.

Second, we investigate the influence of the threshold values on our three loss parametrizations.
For the \emph{first loss parametrization}, we study the influence of $\threshold_h$ which conditions the proportion of false positive and false negative proposals introduced in the pseudo-labeled dataset.
The performance of the teacher and student models for the different sizes of labeled dataset and for values of $\threshold_h$ ranging from $0.5$ to $0.99$ are shown in \Figure{p1_ablation}.
For all sizes, increasing the threshold value tends to increase the performance.
Furthermore, all student models achieve better performance than the teacher for high threshold values, indicating that even with a simple strategy it is possible to improve on supervised methods using unlabeled data.
For the student model trained with $5\%$ and $10\%$ of the labeled dataset, the optimal threshold value corresponds to $\threshold_h=0.99$, showing that it is better to be more selective at the expense of generating false negatives, rather than introducing false positives in the unlabeled dataset.
%For the model trained with $1\%$, we can see that the best threshold is $\threshold_h=0.9$ as otherwise too few positive samples are selected.
% To understand the difference between those results, we can analyze the histograms of score predictions for the labeled sizes of $1\%$ and $10\%$, which are displayed in Figure~\ref{fig:histo_comparison}.

% \begin{figure*}
%     \centering
%     \begin{subfigure}{.5\textwidth}
%       \centering
%       \includegraphics[width=\linewidth]{figures/histo_1_v2.pdf}
%       \label{fig:histo_1}
%     \end{subfigure}%
%     \begin{subfigure}{.5\textwidth}
%       \centering
%       \includegraphics[width=\linewidth]{figures/histo_10_v2.pdf}
%       \label{fig:histo_10}
%     \end{subfigure}
%     \caption{Histogram}
%     \label{fig:histo_comparison}
% \end{figure*}

\input{tables/table4}
\input{tables/table2}
\begin{figure*}
    \centering
    \includegraphics[width=\textwidth]{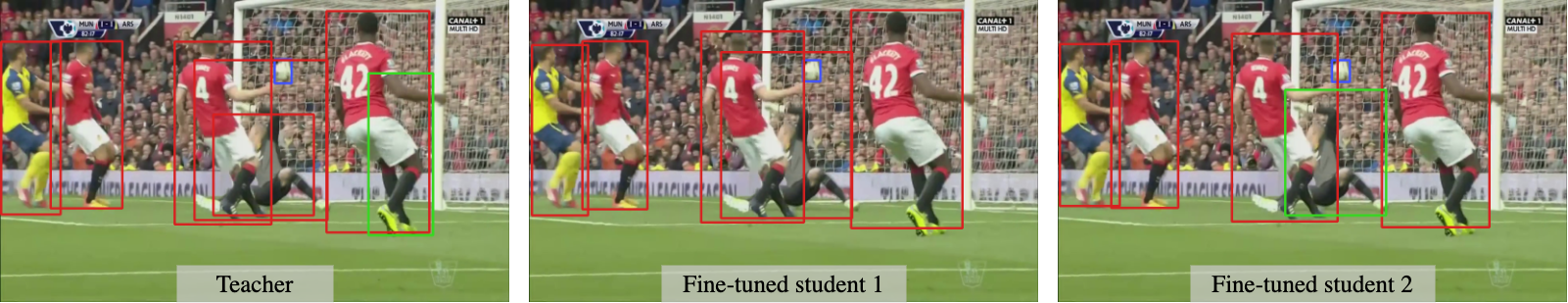}
    \caption{\textbf{Qualitative results.} Comparison of the detections on a test set image for the first teacher (left), fine-tuned student model after $1$ iteration (middle), and fine-tuned student model after $2$ iterations (right). The considered labeled dataset size is $10\%$, with $10$ extra unlabeled games, using the third parametrization for both iterations, with the optimal threshold values presented in Table~\ref{tab:main-results}.}
    \label{fig:qualitative}
\end{figure*}
For the \emph{second parametrization}, we analyze the influence of $\threshold_l$ and $\threshold_h$ independently, and provide the results only on $10\%$ of the labeled dataset with $10$ extra unlabeled games, since the other labeled dataset sizes lead to similar observations.
Our setup is the following: (1) we vary $\threshold_h$ from $0.6$ to $0.9$ with a fixed value of $\threshold_l=0.5$, and (2) we vary $\threshold_l$ from $0.6$ to $0.8$ with a fixed value of $\threshold_h=0.9$.
We also evaluate this parametrization with higher threshold values ($\threshold_l=0.99$ and $\threshold_h=0.999$).
All results are presented in Table~\ref{tab:p2_table}.
Similarly to the first parametrization, we see that the performance increases with $\threshold_h$.
In addition, higher values for $\threshold_l$ also lead to better performance.
This means that the transition zone between true negatives and positives is around high confidence scores.
In other words, detected objects with confidence scores lower than $0.8$ should be considered as negative samples rather than being ignored.
By construction, this observation is dependent on the considered network architecture and dataset. 
However, it provides good insights on how we should consider the Faster R-CNN predictions based on their confidence scores.
We can also observe that a very high value for $\threshold_l$ and $\threshold_h$ reduces the performance of the student.%, as for the first parametrization on $1\%$ of the labeled data.

For the \emph{third parametrization}, we also study the influence of $\threshold_l$ and $\threshold_h$ on the performance.
From our experiments, we noticed that the best performance is always obtained when choosing $\threshold_h=1$.
This means that we should increasingly give credit to the predictions based on their confidence score with no upper limit, independently of the value of $\threshold_l$.
Therefore, we show the performance when varying $\threshold_l$ only for this optimal threshold ($\threshold_h=1$).
As can be seen in Table~\ref{tab:p3_table}, the performance increases with the value of $\threshold_l$, showing that we should consider predictions with a higher prediction score than before ($\threshold_l=0.9$).
In fact, the predictions between the thresholds are not completely ignored compared to the second parametrization, but are simply less considered when approaching $\threshold_l$.
\input{tables/table3}

Finally, since our method may also be used in an iterative fashion, we provide some insights on to what extend a second iteration of pseudo-labelling using the first student as the new teacher and training a second student further improve the performance.
In particular, we study the iterative process with $10\%$ of the labeled dataset and the third parametrization since it gives good performance for one iteration and that its training time is reasonable.
As mentioned earlier in Table~\ref{tab:main-results}, for this setup, the first teacher and the first fine-tuned student have performances of $39.5\%$ and $43.7\%$, respectively.
After fine-tuning, the second student model reaches an mAP of $45.1\%$, which further increases the performance compared to the teacher and the first student.
%However, the relative gain is smaller than from the first iteration.
In further work, we will study more deeply our iterative process, especially when considering the whole labeled and unlabeled dataset, which is computationally intensive.

\mysection{Qualitative results.}
Illustrations of our method's predictions for consecutive iterations are shown in Figure~\ref{fig:qualitative} for the first teacher, the first student, and the second student.
As can be seen, the first student does not produce false positives, unlike the teacher, but fails at correctly localizing and classifying the goalkeeper.
However, the second student manages to correctly detect the goalkeeper.
This perfectly illustrates the detection improvements at each iteration.

%% file: tables/table1.tex
\begin{table}
\scriptsize
    \caption{\textbf{Best performances of the teacher and the fine-tuned student after a single iteration.} Performance of our method are given for several labeled dataset sizes, trained with a fixed amount of $10$ extra unlabeled games (that is $55{,}000$ frames). According to best practices, hyper parameters such as the threshold values of our parametric losses are optimized on the validation set only. In addition, the performances for the test set are calculated after training with the entire labeled and unlabeled datasets, and the optimal parameters obtained on the validation set.
    The mAP value of $\mathbf{52.3\%}$ is the first detection benchmark
on the new SoccerNet-v3 dataset.  ($^{\dagger}$corresponds to $\threshold_h=0.9$)}
    % \setlength{\tabcolsep}{2.5pt}
    % \tabcolsep{6pt}
    \resizebox{\linewidth}{!}{
    \begin{tabular}{l|c|c||c c c c|| c}
    \toprule
        & & & \multicolumn{4}{c||}{Validation set} & Test set \\
        Method & $\threshold_l$ & $\threshold_h$ & $1\%$ & $5\%$ & $10\%$ & $100\%$ & $100\%$ \\
        \midrule
        Teacher & - & - & $18.1$ & $31.9$ & $39.5$ & $52.7$ & $51.0$ \\
        Param. $1$ & - & $0.99$ & $25.8^{\dagger}$ & $38.6$ & $\mathbf{44.3}$ & $53.7$ & $-$ \\
        Param. $2$ & $0.9$ & $0.99$ & $26.0$ & $38.7$ & $\mathbf{44.3}$ & $\mathbf{53.8}$ & $-$\\
        Param. $3$ & $0.9$ & $1$ & $\mathbf{26.2}$ & $\mathbf{38.9}$ & $43.7$ & $\mathbf{53.8}$ & $\mathbf{52.3}$ \\
    \bottomrule
    \end{tabular}%
    }
    \label{tab:main-results}
\end{table}

%% file: tables/table4.tex
\begin{table}
\scriptsize
    \caption{\textbf{Fine-tuning comparison.} Performance improvement when fine-tuning the student network on the labeled dataset at the end of the training for different labeled dataset sizes, with $10$ extra unlabeled games. After fine-tuning, the performance increase for all dataset sizes and all parametrizations, showing the importance of this last training step ($^{\dagger}$corresponds to $\threshold_h=0.9$).
    }
    \centering
    % \setlength{\tabcolsep}{2.5pt}
    % \tabcolsep{6pt}
    \resizebox{\linewidth}{!}{
    \begin{tabular}{l|c c c c}
    \toprule
        Method & $1\%$ & $5\%$ & $10\%$ & $100\%$ \\
        \midrule
        Teacher & $18.1$ & $31.9$ & $39.5$ & $52.7$ \\
        Param. $1$ & $22.6^{\dagger} \rightarrow 25.8$ & $36.0 \rightarrow 38.6$ & $42.3 \rightarrow 44.3$ & $52.6 \rightarrow 53.7$ \\
        Param. $2$ & $23.1 \rightarrow 26.1$ & $36.6 \rightarrow 38.7$ & $43.0 \rightarrow 44.3$ & $52.6 \rightarrow 53.8$ \\
        Param. $3$ & $23.0 \rightarrow 26.2$ & $36.1 \rightarrow 38.9$ & $41.9 \rightarrow 43.7$ & $52.7 \rightarrow 53.8$ \\
    \bottomrule
    \end{tabular}}
    \label{tab:finetuning}
\end{table}

%% file: tables/table2.tex
% \begin{table}[ht]
% \scriptsize
%     \caption{}
%     \centering
%     \vspace{0.3cm}
%     % \setlength{\tabcolsep}{2.5pt}
%     % \tabcolsep{6pt}
%     % \resizebox{\linewidth}{!}{
%     \begin{tabular}{c|c|c}
%         $\threshold_l$ & $\threshold_h$ & mAP \\
%         \midrule
%         $0.5$ & $0.6$ & $38.12$ \\
%         $0.5$ & $0.7$ & $38.97$ \\
%         $0.5$ & $0.8$ & $39.48$ \\
%         $0.5$ & $0.9$ & $40.36$ \\
%         \midrule
%         $0.6$ & $0.9$ & $40.86$ \\
%         $0.7$ & $0.9$ & $41.15$ \\
%         $0.8$ & $0.9$ & $41.35$ \\
%         \midrule
%         $0.9$ & $0.99$ & $\mathbf{43.03}$ \\
%     \end{tabular}%}
%     \label{tab:p2_table}
% \end{table}

\begin{table}
\footnotesize
    \caption{\textbf{Optimal threshold values for the second parametrization.} Comparison of the performance of the second parametrization before fine-tuning for different threshold values $\threshold_l$ and $\threshold_h$ on $10\%$ of the labeled dataset size with $10$ extra games as unlabeled data. The performance of the student increases with both threshold values, indicating that predictions should be considered as background samples for high values of the confidence score as well.}
    \resizebox{\linewidth}{!}{
    \begin{tabular}{c|ccccccccc}
        \toprule
        $\threshold_l$ & $0.5$ & $0.5$ & $0.5$ & $0.5$ & $0.6$ & $0.7$ & $0.8$ & $0.9$ & $0.99$ \\
        \midrule
        $\threshold_h$ & $0.6$ & $0.7$ & $0.8$ & $0.9$ & $0.9$ & $0.9$ & $0.9$ & $0.99$ & $0.999$ \\
        \midrule
        mAP & $38.1$ & $39.0$ & $39.5$ & $40.4$ & $40.9$ & $41.1$ & $41.4$ & $\mathbf{43.0}$ & $41.0$ \\
        \bottomrule
    \end{tabular}}
    \label{tab:p2_table}
\end{table}

%% file: tables/table3.tex
\begin{table}
\footnotesize
    \caption{\textbf{Optimal threshold values for the third parametrization.} Comparison of the performance of the third parametrization before fine-tuning for different threshold values $\threshold_l$ when $\threshold_h=1$, on $10\%$ of the labeled data and $10$ extra games as unlabeled data. The performance of the student increases with $\threshold_l$ showing that only high confidence samples should be considered.}
    % \setlength{\tabcolsep}{2.5pt}
    % \tabcolsep{6pt}
    % \resizebox{\linewidth}{!}{
    \centering
    \begin{tabular}{c|ccccc}
        \toprule
        $\threshold_l$ & $0.5$ & $0.6$ & $0.7$ & $0.8$ & $0.9$ \\
%        \midrule
%        $\threshold_h$ & $1$ & $1$ & $1$ & $1$ & $1$ \\
        \midrule
        mAP & $39.1$ & $40.1$ & $40.8$ & $41.3$ & $\mathbf{41.9}$\\
        \bottomrule
    \end{tabular}%}
    \label{tab:p3_table}
\end{table}

%% file: sections/5_conclusion.tex
\section{Conclusion}
\label{sec:Conclusion}

In this work, we propose a new generic semi-supervised method based on a teacher-student approach for object detection. In particular, we show how unlabeled data improves the detection performance of a model trained solely on labeled data. Our method consists in using a teacher trained on labeled data to produce surrogate ground-truth annotations on the unlabeled dataset, later added to the labeled data to train a student model. To adapt the training process to our scenario, we propose three loss parametrizations based on the confidence score of the teacher's predictions to introduce doubt. By doing so, our method substantially improves the performance compared to  supervised training. A side result is that we set the first detection benchmark on the new SoccerNet-v3 dataset. 
Since our method is data and network agnostic, we presume that it is always possible to use available unlabeled data, a common situation in sports analysis, to further improve a detection network. %This might well be an important milestone in the development of sport analysis.
% As future work, we will further explore the iterative process, not limiting our study to two iterations and a small labeled dataset. We will also study how to reduce the training time, which is costly at the moment and prevents us from doing multiple iterations on the whole dataset.

%% file: ms.bbl
\begin{thebibliography}{10}\itemsep=-1pt

\bibitem{Sangesa2020UsingPB}
Adri{\`a} Arbu{\'e}s~Sang{\"u}esa, Adri{\`a}n Mart{\'i}n, Javier Fern{\'a}ndez,
  Coloma Ballester, and Gloria Haro.
\newblock Using player's body-orientation to model pass feasibility in soccer.
\newblock In {\em IEEE Int. Conf. Comput. Vis. and Pattern Recogn. Workshops
  (CVPRW)}, pages 3875--3884, Seattle, WA, USA, June 2020.

\bibitem{Archana2015AnEfficient}
M. Archana and M. Geetha.
\newblock An efficient ball and player detection in broadcast tennis video.
\newblock In {\em Intelligent Systems Technologies and Applications}, volume
  384 of {\em Adv. in Intell. Syst. and Comput.}, pages 427--436. Springer,
  2015.

\bibitem{Berthelot2020ReMixMatch}
David Berthelot, Nicholas Carlini, Ekin Cubuk, Alex Kurakin, Kihyuk Sohn, Han
  Zhang, and Colin Raffel.
\newblock Re{M}ix{M}atch: Semi-supervised learning with distribution matching
  and augmentation anchoring.
\newblock In {\em Int. Conf. on Learn. Rep. (ICLR)}, Addis Abada, Ethiopia,
  Apr.-May 2020.

\bibitem{Berthelot2019MixMatch}
David Berthelot, Nicholas Carlini, Ian Goodfellow, Nicolas Papernot, Avital
  Oliver, and Colin Raffel.
\newblock Mix{M}atch: A holistic approach to semi-supervised learning.
\newblock In {\em Adv. in Neural Inform. Process. Syst. (NeurIPS)}, volume~32,
  Vancouver, Canada, Dec. 2019. Curran Associates, Inc.

\bibitem{Cioppa2022SoccerNetv3}
Anthony Cioppa, Adrien Deli{\`e}ge, Silvio Giancola, Bernard Ghanem, and Marc
  Van~Droogenbroeck.
\newblock {SoccerNet-v3}: Scaling up soccernet with multi-view spatial
  localization and re-identification.
\newblock {\em Submitted to Scientific Data}, 2022.

\bibitem{Cioppa2021Camera}
Anthony Cioppa, Adrien Deli{\`e}ge, Silvio Giancola, Floriane Magera, Olivier
  Barnich, Bernard Ghanem, and Marc Van~Droogenbroeck.
\newblock Camera calibration and player localization in {SoccerNet-v2} and
  investigation of their representations for action spotting.
\newblock In {\em IEEE Int. Conf. Comput. Vis. and Pattern Recogn. Workshops
  (CVPRW), CVsports}, pages 4537--4546, Nashville, TN, USA, June 2021.

\bibitem{Cioppa2020Multimodal}
Anthony Cioppa, Adrien Deli{\`e}ge, Noor Ul~Huda, Rikke Gade, Marc
  Van~Droogenbroeck, and Thomas~B. Moeslund.
\newblock Multimodal and multiview distillation for real-time player detection
  on a football field.
\newblock In {\em IEEE Int. Conf. Comput. Vis. and Pattern Recogn. Workshops
  (CVPRW), CVsports}, pages 3846--3855, Seattle, WA, USA, June 2020.

\bibitem{Cioppa2019ARTHuS}
Anthony Cioppa, Adrien Deliège, Maxime Istasse, Christophe De~Vleeschouwer,
  and Marc Van~Droogenbroeck.
\newblock {ARTHuS}: Adaptive real-time human segmentation in sports through
  online distillation.
\newblock In {\em IEEE Int. Conf. Comput. Vis. and Pattern Recogn. Workshops
  (CVPRW), CVsports}, pages 2505--2514, Long Beach, CA, USA, June 2019.

\bibitem{Deliege2021SoccerNetv2}
Adrien Deli{\`e}ge, Anthony Cioppa, Silvio Giancola, Meisam~J. Seikavandi,
  Jacob~V. Dueholm, Kamal Nasrollahi, Bernard Ghanem, Thomas~B. Moeslund, and
  Marc Van~Droogenbroeck.
\newblock {SoccerNet}-v2: A dataset and benchmarks for holistic understanding
  of broadcast soccer videos.
\newblock In {\em IEEE Int. Conf. Comput. Vis. and Pattern Recogn. Workshops
  (CVPRW), CVsports}, pages 4508--4519, Nashville, TN, USA, June 2021.
\newblock Best CVSports paper award.

\bibitem{Everingham2010PascalVOC}
Mark Everingham, Luc Van~Gool, Christopher K.~I. Williams, John Winn, and
  Andrew Zisserman.
\newblock The {PASCAL} visual object classes {(VOC)} challenge.
\newblock {\em Int. J. Comp. Vis.}, 88(2):303--338, June 2010.

\bibitem{Giancola2018SoccerNet}
Silvio Giancola, Mohieddine Amine, Tarek Dghaily, and Bernard Ghanem.
\newblock {SoccerNet}: A scalable dataset for action spotting in soccer videos.
\newblock In {\em IEEE Int. Conf. Comput. Vis. and Pattern Recogn. Workshops
  (CVPRW)}, pages 1711--1721, Salt Lake City, UT, USA, June 2018.

\bibitem{Girshick2015Fast}
Ross Girshick.
\newblock Fast {R-CNN}.
\newblock In {\em IEEE Int. Conf. Comput. Vis. (ICCV)}, pages 1440--1448,
  Santiago, Chile, Dec. 2015.

\bibitem{Girshick2014Rich}
Ross Girshick, Jeff Donahue, Trevor Darrell, and Jitendra Malik.
\newblock Rich feature hierarchies for accurate object detection and semantic
  segmentation.
\newblock In {\em IEEE Int. Conf. Comput. Vis. and Pattern Recogn. (CVPR)},
  pages 580--587, Columbus, OH, USA, June 2014.

\bibitem{Gough2021Market}
Christina Gough.
\newblock Market size of the sports analytics industry worldwide in 2020 and
  2028, 2021.
  \url{https://www.statista.com/statistics/1185536/sports-analytics-market-size}.

\bibitem{He2017Mask}
Kaiming He, Georgia Gkioxari, Piotr Dollár, and Ross Girshick.
\newblock Mask {R-CNN}.
\newblock In {\em IEEE Int. Conf. Comput. Vis. (ICCV)}, pages 2980--2988,
  Venice, Italy, Oct. 2017.

\bibitem{He2016DeepResidual}
Kaiming He, Xiangyu Zhang, Shaoqing Ren, and Jian Sun.
\newblock Deep residual learning for image recognition.
\newblock In {\em IEEE Int. Conf. Comput. Vis. and Pattern Recogn. (CVPR)},
  pages 770--778, Las Vegas, NV, USA, June 2016.

\bibitem{Hurault2020SelfSupervised}
Samuel Hurault, Coloma Ballester, and Gloria Haro.
\newblock Self-supervised small soccer player detection and tracking.
\newblock In {\em Int. ACM Workshop Multimedia Content Anal. in Sports
  (MMSports)}, pages 9--18, Seattle, WA, USA, Oct. 2020.

\bibitem{Mordorintelligence2022Sports}
Mordor Intelligence.
\newblock Sports analytics market -- {G}rowth, trends, {COVID}-19 impact, and
  forecasts (2022 - 2027), 2022.
  \url{https://www.mordorintelligence.com/industry-reports/sports-analytics-market}.

\bibitem{Jeong2019Consistency}
Jisoo Jeong, Seungeui Lee, Jeesoo Kim, and Nojun Kwak.
\newblock Consistency-based semi-supervised learning for object detection.
\newblock In {\em Adv. in Neural Inform. Process. Syst. (NeurIPS)}, volume~32,
  Vancouver, Canada, Dec. 2019. Curran Associates, Inc.

\bibitem{Jiang2020SoccerDB}
Yudong Jiang, Kaixu Cui, Leilei Chen, Canjin Wang, and Changliang Xu.
\newblock {SoccerDB}: A large-scale database for comprehensive video
  understanding.
\newblock In {\em Int. ACM Workshop Multimedia Content Anal. in Sports
  (MMSports)}, pages 1--8, 2020.

\bibitem{Kamble2019Deep}
Paresh~R. Kamble, Avinash~G. Keskar, and Kishor~M. Bhurchandi.
\newblock A deep learning ball tracking system in soccer videos.
\newblock {\em Opto-Electronics Review}, 27(1):58--69, Mar. 2019.

\bibitem{DTAI2019WhySports}
DTAI Sports~Analytics Lab.
\newblock Why sports analytics, 2019.
  \url{https://dtai.cs.kuleuven.be/sports/}.

\bibitem{Li2020Improving}
Yandong Li, Di Huang, Danfeng Qin, Liqiang Wang, and Boqing Gong.
\newblock Improving object detection with selective self-supervised
  self-training.
\newblock In {\em Eur. Conf. Comput. Vis. (ECCV)}, volume 12374 of {\em Lect.
  Notes Comp. Sci.}, pages 589--607. Springer, Oct. 2020.

\bibitem{Lin2017Feature}
Tsung-Yi Lin, Piotr Dollar, Ross Girshick, Kaiming He, Bharath Hariharan, and
  Serge Belongie.
\newblock Feature pyramid networks for object detection.
\newblock In {\em IEEE Int. Conf. Comput. Vis. and Pattern Recogn. (CVPR)},
  pages 2117--2125, Honolulu, HI, USA, July 2017.

\bibitem{Lin2017Focal}
Tsung-Yi Lin, Priya Goyal, Ross Girshick, Kaiming He, and Piotr Dollar.
\newblock Focal loss for dense object detection.
\newblock {\em CoRR}, abs/1708.02002, 2017.

\bibitem{Lin2014Microsoft}
Tsung-Yi. Lin, Michael Maire, Serge Belongie, James Hays, Pietro Perona, Deva
  Ramanan, Piotr Doll{\'a}r, and C.~Lawrence Zitnick.
\newblock Microsoft {COCO}: {C}ommon {O}bjects in {C}ontext.
\newblock In {\em Eur. Conf. Comput. Vis. (ECCV)}, volume 8693 of {\em Lect.
  Notes Comp. Sci.}, pages 740--755. Springer, Sept. 2014.

\bibitem{Liu2016SSD}
Wei Liu, Dragomir Anguelov, Dumitru Erhan, Christian Szegedy, Scott Reed,
  Cheng-Yang Fu, and Alexander Berg.
\newblock {SSD}: {S}ingle shot multibox detector.
\newblock {\em CoRR}, abs/1512.02325, 2016.

\bibitem{Liu2021DetectingAM}
Yang Liu, Luiz Hafemann, Michael Jamieson, and Mehrsan Javan.
\newblock Detecting and matching related objects with one proposal multiple
  predictions.
\newblock In {\em IEEE Int. Conf. Comput. Vis. and Pattern Recogn. Workshops
  (CVPRW)}, pages 4515--4522, Nashville, TN, USA, June 2021.

\bibitem{Liu2021Unbiased}
Yen-Cheng Liu, Chih-Yao Ma, Zijian He, Chia-Wen Kuo, Kan Chen, Pzizhao Zhang,
  Bichen Wu, Zsolt Kira, and Peter Vajda.
\newblock Unbiased teacher for semi-supervised object detection.
\newblock In {\em Int. Conf. on Learn. Rep. (ICLR)}, May 2021.

\bibitem{Manafifard2017ASurvey}
Mehrtash Manafifard, Hamid Ebadi, and Hamid Abrishami~Moghaddam.
\newblock A survey on player tracking in soccer videos.
\newblock {\em Comp. Vis. and Image Underst.}, 159:19--46, June 2017.

\bibitem{Oliver2018Realistic}
Avital Oliver, Augustus Odena, Colin Raffel, Ekin~D. Cubuk, and Ian Goodfellow.
\newblock Realistic evaluation of deep semi-supervised learning algorithms.
\newblock In {\em Adv. in Neural Inform. Process. Syst. (NeurIPS)}, volume~31,
  Montr{\'e}al, Canada, Dec. 2018. Curran Associates, Inc.

\bibitem{Pappalardo2019Apublic}
Luca Pappalardo, Paolo Cintia, Alessio Rossi, Emanuele Massucco, Paolo
  Ferragina, Dino Pedreschi, and Fosca Giannotti.
\newblock A public data set of spatio-temporal match events in soccer
  competitions.
\newblock {\em Scientific Data}, 6:1--15, Oct. 2019.

\bibitem{Pham2021Meta}
Hieu Pham, Zihang Dai, QIzhe Xie, Minh-Thang Luong, and Quoc Le.
\newblock Meta pseudo labels.
\newblock In {\em IEEE Int. Conf. Comput. Vis. and Pattern Recogn. (CVPR)},
  pages 11557--11568, Nashville, TN, USA, June 2021.

\bibitem{Pobar2018MaskRCNN}
Miran Pobar and Marina Ivasic-Kos.
\newblock {Mask R-CNN} and optical flow based method for detection and marking
  of handball actions.
\newblock In {\em Int. Congress on Image and Signal Process., BioMedical Eng.
  and Inform. (CISP-BMEI)}, pages 1--6, Beijing, China, Oct. 2018.

\bibitem{Rao2015ANovel}
Upendra~M. Rao and Umesh~C. Pati.
\newblock A novel algorithm for detection of soccer ball and player.
\newblock In {\em Int. Conf. Commun. and Signal Process. (ICCSP)}, pages
  344--348, Melmaruvathur, India, Apr. 2015.

\bibitem{Redmon2016YOLO}
Joseph Redmon, Santosh Divvala, Ross Girshick, and Ali Farhadi.
\newblock You only look once: {U}nified, real-time object detection.
\newblock {\em CoRR}, abs/1506.02640, June 2016.

\bibitem{Redmon2018YOLOv3}
Joseph Redmon and Ali Farhadi.
\newblock {YOLO}v3: An incremental improvement.
\newblock {\em CoRR}, abs/1804.02767, Apr. 2018.

\bibitem{Ren2017Faster}
Shaoqing Ren, Kaiming He, Ross Girshick, and Jian Sun.
\newblock {Faster R-CNN}: Towards real-time object detection with region
  proposal networks.
\newblock {\em IEEE Trans. Pattern Anal. Mach. Intell.}, 39(6):1137--1149, June
  2017.

\bibitem{Sah2018Evaluation}
Melike Sah and Cem Direkoglu.
\newblock Evaluation of image representations for player detection in field
  sports using convolutional neural networks.
\newblock In {\em International Conference on Theory and Application of Fuzzy
  Systems and Soft Computing (ICAFS)}, volume 896 of {\em Adv. in Intell. Syst.
  and Comput.}, pages 107--115. Springer, 2018.

\bibitem{Sohn2020FixMatch}
Kihyuk Sohn, David Berthelot, Chun-Liang Li, Zizhao Zhang, Nicholas Carlini,
  Ekin~D. Cubuk, Han Kurakin, Alexand~Zhang, and Colin Raffel.
\newblock Fix{M}atch: Simplifying semi-supervised learning with consistency and
  confidence.
\newblock In {\em Adv. in Neural Inform. Process. Syst. (NeurIPS)}, volume~33,
  pages 596--608. Curran Associates, Inc., Dec. 2020.

\bibitem{Sohn2020Simple}
Kihyuk Sohn, Zizhao Zhang, Chun-Liang Li, Han Zhang, Chen-Yu Lee, and Tomas
  Pfister.
\newblock A simple semi-supervised learning framework for object detection.
\newblock {\em CoRR}, abs/2005.04757, 2020.

\bibitem{Tan2020EfficientDet}
Mingxing Tan, Ruoming Pang, and Quoc~V. Le.
\newblock Efficient{D}et: Scalable and efficient object detection.
\newblock In {\em IEEE Int. Conf. Comput. Vis. and Pattern Recogn. (CVPR)},
  pages 10778--10787, Seattle, WA, USA, June 2020.

\bibitem{Tang2021Proposal}
Peng Tang, Chetan Ramaiah, Yan Wang, Ran Xu, and Caiming Xiong.
\newblock Proposal learning for semi-supervised object detection.
\newblock In {\em IEEE Winter Conf. Applicat. Comp. Vis. (WACV)}, pages
  2291--2301, Waikoloa, HI, USA, Jan. 2021.

\bibitem{Thomas2017Computer}
Graham Thomas, Rikke Gade, Thomas~B. Moeslund, Peter Carr, and Adrian Hilton.
\newblock Computer vision for sports: current applications and research topics.
\newblock {\em Comp. Vis. and Image Underst.}, 159:3--18, June 2017.

\bibitem{Xie2020SelfTraining}
Qizhe Xie, Minh-Thang Luong, Eduard Hovy, and Quoc~V. Le.
\newblock Self-training with noisy student improves {ImageNet} classification.
\newblock In {\em IEEE Int. Conf. Comput. Vis. and Pattern Recogn. (CVPR)},
  pages 10684--10695, Seattle, WA, USA, June 2020.

\bibitem{Xu2021Soft}
Mengde Xu, Zheng Zhang, Han Hu, Jianfeng Wang, Lijuan Wang, Fangyun Wei, Xiang
  Bai, and Zicheng Liu.
\newblock End-to-end semi-supervised object detection with soft teacher.
\newblock In {\em IEEE Int. Conf. Comput. Vis. (ICCV)}, pages 3060--3069,
  Montr{\'e}al, Canada, Oct. 2021.

\bibitem{Yang20183DMultiview}
Yukun Yang, Min Xu, Wanneng Wu, Ruiheng Zhang, and Yu Peng.
\newblock {3D} multiview basketball players detection and localization based on
  probabilistic occupancy.
\newblock In {\em Digit. Image Comp.: Tech. and Applicat.}, pages 1--8,
  Canberra, ACT, Australia, Dec. 2018.

\bibitem{Yu2018Comprehensive}
Junqing Yu, Aiping Lei, Zikai Song, Tingting Wang, Hengyou Cai, and Na Feng.
\newblock Comprehensive dataset of broadcast soccer videos.
\newblock In {\em IEEE Conf. on Multimedia Inform. Process. and Retrieval
  (MIPR)}, pages 418--423, Miami, FL, USA, June 2018.

\bibitem{Zoph2020Rethinking}
Barret Zoph, Golnaz Ghiasi, Tsung-Yi Lin, Yin Cui, Hanxiao Liu, Ekin~D. Cubuk,
  and Quoc~V. Le.
\newblock Rethinking pre-training and self-training.
\newblock In {\em Adv. in Neural Inform. Process. Syst. (NeurIPS)}, volume~33,
  pages 3833--3845. Curran Associates, Inc., Dec. 2020.

\end{thebibliography}
